# Formal Definition of AI

## Dimiter Dobrev
dobrev@2-box.net

18 October, 2005

**Abstract**

A definition of Artificial Intelligence was proposed in [1] but this definition was not absolutely formal at least because the word "Human" was used. In this paper we will formalize the definition from [1]. The biggest problem in this definition was that the level of intelligence of AI is compared to the intelligence of a human being. In order to change this we will introduce some parameters to which AI will depend. One of this parameters will be the level of intelligence and we will define one AI to each level of intelligence. We assume that for some level of intelligence the respective AI will be more intelligent than a human being. Nevertheless, we cannot say which is this level because we cannot calculate its exact value.

## Introduction

The definition in [1] first was published in popular form in [2, 3]. It was stated in one sentence but with many assumptions and explanations which were given before and after this sentence. Here is the definition of AI in one sentence:

**AI will be such a program which in an arbitrary world will cope no worse than a human.**

From this sentence you can see that we assume that AI is a program. Also, we assume that AI is a step device and that on every step it inputs from outside a portion of information (a letter from finite alphabet $\Sigma$) and outputs a portion of information (a letter from a finite alphabet $\Omega$). The third assumption is that AI is in some environment which gives it a portion of information on every step and which receives the output of AI. Also, we assume that this environment will be influenced of the information which AI outputs. This environment can be natural or artificial and we will refer to it as "World".

The World will be: one set S, one element $s_0$ of S and two functions World(s, d) and View(s). The set S contains the internal states of the world and it can be finite or infinite. The element $s_0$ of S will be the world's starting state. The function World will take as arguments the current state of the world and the influence that our device exerts on the world at the current step. As a result, this function will return the new state of the world (which it will obtain on the next step). The function View gives the information what our device sees. An argument of this function will be the world's current state and the returned value will be the information that the device will receive (at a given step).

Life in one world will be any infinite row of the type: $d_1, v_1, d_2, v_2, ...$ where $v_i$ are letters from $\Sigma$ and $d_i$ are letters from $\Omega$. Also, there has to exist infinite row $s_0, s_1, s_2, ...$ such that $s_0$ is the staring state of the world and $\forall i > 0$ $v_i = \text{View}(s_i)$ and $\forall i$ $s_{i+1} = \text{World}(s_i, d_{i+1})$. It is obvious that if the world is given then the life depends only on the actions of AI (i.e. depends only on the row $d_1, d_2, d_3, ...$).

In order to transform the definition in [1] and to make it formal, we have to define what is a program, what is a good world and when one life is better than another.

The first task is easy because this work is done by Turing in the main part. Anyway, the Turing definition of program is for a program which represents function, but here we need a transducer which inputs the row $v_1, v_2, v_3, ...$ and outputs the row $d_1, d_2, d_3, ...$ . So, we will make a modification of the definition of Turing machine [9].

Our second task is to say what is a good world. It was written in [1] that if you can make a fatal error in one world then this world is not good. What is world without fatal errors needs additional formalization.

The next problem is to say when one life is better than another. This is done in [1] but there are some problems connected with the infinity which have to be fixed.

The last task is to say how intelligent our program should be and this cannot be done by comparison with a human being.

## What is a program

We will define a program as a Turing machine [9]. Let its alphabet $\Delta$ consist of the letters from $\Sigma$, from $\Omega$, from one blank symbol $\lambda$ and from some service signs.

Let our Turing machine have finite set of internal states $P$, one starting state $p_0$ and a partial function $F : P \times \Delta \to P \times \Delta \times \{\text{Left, Right}\}$.

The Turing machine (TM) is a step device and it makes steps in order to do calculations. On the other hand, AI is a step device and its life consists of steps. In order to make distinction between these two types of steps we will call them small and big steps. When we speak about time we will mean the number of big steps.

Of course, our TM will start from the state $p_0$ with infinite tape filled with the blank symbol $\lambda$. How our TM will make one small step. If it is in state $p$ and if its head looks at the letter $\delta$ then $F(p, \delta)$ will be a 3-tuple which first element is the new state after the small step, the second element will be the new letter which will replace δ on the tape and the third element will be direction in which the head will move.

How will our TM (which is also our AI) make one big step? This will happen when after a small step the new state of TM is again $p_0$. At this moment our TM has to output one letter $d_i$ and to input one letter $v_i$. We will assume that the letter which is outputted is that which is written on the place of δ on this small step. But how after outputting the letter $d_i$ will our TM input the letter $v_i$? We will put this letter on the place where the head after the small step is. In this way we are intervening in the work of the TM by replacing one symbol from the tape with another. The replaced symbol is lost in some sense because it will not influence the execution of the TM from this small step on.

We will assume that our TM is outputting only letters from Ω (no letters from the rest of Δ). Also, we assume that our TM never hangs. TM hangs if after reading some input $v_1$, $v_2$, ... , $v_n$ it stops because it falls into some state p and its head falls on some letter δ such that F(p, δ) is not defined. TM also hangs if after reading of some input $v_1$, $v_2$, ... , $v_n$ it makes infinitely many small steps without reaching the state $p_0$ (without making of big steps anymore).

After this we have a formal definition of a program. We have to mention that there is no restriction on the number of the small steps which TM needs to make for one big step. This number has to be finite but it is not restricted. Maybe it is a good idea to add one parameter Max_number_of_small_steps_in_AI in order to exclude some decisions for AI which are combinatory explosions. (If we restrict the number of small steps then we have to restrict also the number of service signs in Δ because we can speed up the TM by increasing the size of its alphabet.) If we want to use AI as a real program on a real computer then we have to take into consideration that the memory of the real computers is limited. So, we can restrict also the size of the tape. Anyway, we will not care about the efficiency of AI and we will not make such restrictions.

## What is a world without fatal errors

It is very difficult to define what is a world without fatal errors. That is why we will do something else. We will restrict our set of worlds is such a way that the new set will contain only worlds without fatal errors.

Let our world look like one infinite sequence of games. Let every game be independent from the previous ones. Let us have three special letters in Σ, which we will call final letters. Let this letters be {victory, loss, draw}. Let every game finish with one of the final letters. Let every game be shorter than 1000 big steps.

**Remark 1**: Our definition of AI will depend on many parameters. In order to simplify the exposition we will fix these parameters to concrete numbers. Such parameter is the maximum number of steps in a game which will be fixed to 1000. Also, in order to simplify the exposition we will use different numbers for different parameters.

**Remark 2**: The only parameters in our definition which are not numbers are the alphabets Σ and Ω. We will assume that these alphabets are fixed and that Ω has at least 2 letters and Σ has at least 2 letters which are not final. (If Ω has only one letter then there will be no choice for the action of AI and the world will be independent from this action. If Σ has only one letter, which is not final, then the game will be blind because AI will not receive any information until the end of the game. Therefore, the minimum for |Σ| is 5.)

We will assume that the world has three special internal states {s_victory, s_loss, s_draw}, which we will call final states. Let these states be indistinguishable from the state $s_0$ for the function World. This means that the world will behave in the final states in the same way as if it was in the starting state. Let the function View distinguish the final states and return from them the letters victory, loss and draw respectively. Also, the final states will be the only states on which the function View will return one of the letters {victory, loss, draw}.

After the restriction of the definition of World, we can be sure that there are no fatal errors in our world because the life in such a world is an infinite sequence of games and if we lose some games (finitely many) then this will not be fatal because every new game is independent from the previous ones. Also, we are sure that a new game will come sooner or later because every game is finite (i.e. previous game is shorter than 1000 steps).

## When is one life better than another

In [1] we gave the following definition for the meaning of the life: One life is better than another if it includes more good letters and fewer bad letters. Here good letters will be {victory, draw} and bad letters will be {loss, draw}. So, here life is good if we win often and lose seldom.

We want to introduce one function Success which will evaluate with a real number every life in order to say how good it is. For that we will define first the function Success for the every beginning of life (all beginnings are finite). After that we will calculate the limit of Success when the size of the beginnings goes to infinity and this limit will be the value of Success for the entire life.

The function Success can be defined for the beginnings like the difference between the number of victories and the number of losses. This is not a good idea because then the limit of Success will possibly converge to infinity (plus or minus infinity). It is a better idea to calculate the percentage of victories. So, we define Success as (2.N_victory +N_draw)/ (2.N_games). Here N_victory is the number of victories (analogically for N_draw and N_games). Function Success will give us a number between 0 and 1 for every beginning and its limit will be also between 0 and 1. The only problem is that Success may not have a limit. In such a case we will use the average between limit inferior and limit superior.

## Trivial decisions

Now we have a really formal definition of AI and this gives us the first trivial decision for AI.

TD1 will be the program which plays at random until the first victory. After that TD1 repeats this victory forever. For this TD1 needs only to remember what it did in the last game. If the last game was victorious then it can repeat this last game because the function World is deterministic and if TD1 is doing the same then the world will do the same too.

TD1 is perfect in all worlds in which the victory is possible. If the victory is not possible then TD1 will play at random forever. That is why we will make TD2 which will be perfect in all worlds.

TD2 will be this program which tries sequentially all possible game's strategies until it finds the first victory and after that repeats this victory forever. If there is no victorious game strategy then TD2 repeats the last draw game forever. If the draw game is not possible too then TD2 plays at random. (It is important that the game's length is not more than 1000. This means that the number of the game's strategies is finite.)

TD2 is perfect in all worlds and this means that it is a trivial decision on our definition for AI. Really, the definition stated that AI has to cope no worse than a human but for the perfect program this is true because it copes no worse than anything even no worse than a human being.

It is suspicious that such simple program like TD2 can satisfy our definition for AI. That is why we will change the definition by accepting more possible worlds. It is too restrictive to assume that the game is deterministic and every time you do the same the same will happen.

## Nondeterministic games

We will assume that the function World is not deterministic. It is better to say that it is multy-valued function, which chooses at random one of its possible values. Let every possible value correspond to one real number, which is the possibility for this value to be chosen. We will assume also that $\forall s\ \forall \omega$ World(s, $\omega$) has at least one value and that $\forall s\ \forall \omega$ (for every two different values of World(s, $\omega$) the function View returns different result).

**Remark 3**: The latter means that if something nondeterministic happens this information will be given to AI immediately by the function View. There is no sense to assume existence of a nondeterministic change which cannot be detected immediately but later or even which cannot be detected never.

Now we will ask the question what is the best strategy in such a world and we will offer a program, which will be almost perfect. Before that we need several definitions:

**Definition 1: Tree of any game.** It will have two types of vertices. The root and the other vertices which are on even depth will be the vertices of type AI (because they correspond to the moments when AI has to do its choice). The vertices which are on odd depth will be vertices of the type world (because they correspond to the moments when the world will answer at random). From the vertices of type AI go out $|\Omega|$ arcs and to every such arc corresponds one of the letters from $\Omega$. There is one exception. If the arc which is right before this vertex corresponds to a final letter, then this vertex is a leaf. From the vertices of type world go out $|\Sigma|$ arcs and to every such arc corresponds one of the letters from $\Sigma$. Here there is an exception again. If this vertex is on depth 1999, then only three arcs go out and these three arcs correspond to the final letters.

You can see that the tree of any game is finite and its maximum depth is 2000 (because games are not longer than 1000 steps). Nevertheless, there are leaves on any even depth between 2 and 2000.

**Definition 2: Tree of any 100 games.** Let us take the tree of any game. Let us replace all of its leaves with the tree on any game. Let us repeat this operation 99 times. The result will be the tree of any 100 games (which is 100 times deeper than the tree of any game).

From the tree of any game we will receive Strategy for any game. This will be its subtree which is obtained by choosing one vertex from the successors of every vertex of the type AI and deleting the rest successors (and their subtrees). Analogically we make Strategy for any 100 games like a subtree from the tree of any 100 games. We have to mention that the strategy for 100 games can be different from repeating one Strategy for any game 100 times. The reason is because the strategy on the next game can depend on the previous games.

**Definition 3: Tree of this game.** For every concrete game (i.e. concrete world) we can construct the tree of this game as a subtree from the tree of any game. We will juxtapose internal states of the world to the vertices of type AI in the time of this construction. First, we will juxtapose the state $s_0$ to the root. Let $k_0$, $k_1$ and $k_2$ be vertices and let $k_1$ be successor of $k_0$ and $k_2$ be successor of $k_1$. Let $k_0$ be vertex of type AI and let the state s be juxtaposed to it. Let the letters $\omega$ and $\varepsilon$ be juxtaposed to the arcs $<k_0, k_1>$ and $<k_1, k_2>$. In this case if $\varepsilon \neq \text{View}(\text{World}(s, \omega))$ for every value of $\text{World}(s, \omega)$ then we delete the vertex $k_2$ (and its subtree). In the opposite case we juxtapose $k_2$ to this value of $\text{World}(s, \omega)$ for which $\varepsilon = \text{View}(\text{World}(s, \omega))$. This value is only (look at remark 3). Also, we will juxtapose the possibility ε to be the value of View(World(s, ω)) to the arc $\langle k_1, k_2 \rangle$. So, one letter and one possibility will be juxtaposed to the arc $\langle k_1, k_2 \rangle$.

Analogically to the strategy for any game we can make strategy for this game. We have to say that if the World is deterministic (i.e. every vertex of type world has only one successor) then the strategy for this game is a path (a tree without branches). In this case the paths in the tree of this game are exactly the strategies for this game. This was used from TD2 in order to try all strategies.

## Max-Sum algorithm

For every vertex of the tree of this game we can calculate the best possible success (this is our expectation for success, if we play with the best strategy from that vertex on).

1. The best possible success for the leaves will be 1, 0 and 1/2 for the states s_victory, s_loss and s_draw respectively.
2. If the vertex is of type AI, then its best possible success will be the maximum from the best possible successes of its successors.
3. If the vertex is of type world, then its best possible success will be the sum $\sum$ Possibility(i). BestPossibleSuccess(i). Here i runs through all successors of this vertex.

The algorithm for calculating the best possible success can also be used to calculate the best strategy in this game (the best strategy can be more than one). This algorithm looks like the Min-Max algorithm, which we use in chess. Anyway, this is different algorithm, to which we will refer as Max-Sum algorithm. The difference is essential because in the chess we assume that we play against someone who will do the worst thing to us (remark 4). Anyway, in the arbitrary world we cannot assume that the world is against us. For example, when you go to work you go first to the parking lot in order to take your car. If your car is stolen, then you go to the bus stop in order to take the bus. If every time you were presumed the worst case, then you would go directly to the bus stop.

## New trivial decisions

Now we can calculate the best possible success for any game and we will give the next trivial decision (TD3), which will do the best in every game. This means that the success of TD3 for one world will be equal to its best possible success.

TD3 will be the program which plays at random for long time enough. In this time TD3 collects statistical information for the tree of this game and builds inside its memory this tree together with the values of all possibilities. After that time TD3 starts playing by the use of Max-Sum algorithm.

TD3 gives the perfect decision in any world but TD3 is impossible because we cannot say when enough statistical information is selected. Anyway, possible is something which is a little bit worse. For every $\varepsilon > 0$ we will make TD4, which for every world will make success on a distance no more than $\varepsilon$ from the best possible.

TD4 will be this program which simultaneously collects statistical information for the tree of this game and in the same time plays by the use of Max-Sum algorithm on the base of statistics, which is collected up to the current moment. In order to collect statistics TD4 makes experiments which contradict to the recommendations of Max-Sum algorithm. Such experiments are made rarely enough to have success on a distance not bigger than $\varepsilon$ from the best possible success.

We can choose the value of $\varepsilon$ to be as small as we want. Anyway, the price for the small value of $\varepsilon$ is the longer time for education (because of rare experiments). We will call the parameter $\varepsilon$ "courage". Here we receive a surprising conclusion that if AI is more cowardly it is closer to perfection (this is true only in the case of infinite life).

TD4 is a decision for our definition of AI because it is only on $\varepsilon$ distance to perfection unlike the people who are much farther from perfection. We have to mention that in some sense TD4 is not as trivial as TD2, because TD4 represents awful combinatory explosion in the execution time (number of small steps) and in the memory size. Anyway, we said that we will not care about the efficiency of AI for the moment. On the other hand, there is one additional problem, which is present in both TD2 and TD4, and which makes them both useless. This is the problem of the combinatory explosion of the educational time. Imagine that you are playing chess at random against deterministic partner. How long will you need to make accidental victory? Or in case your partner is not deterministic. How long will you need to play all possible game's strategies and try each one several times in order to collect statistical information on how your partner reacts in every case?

## Finite life

In some sense TD2 and TD4 are extremely stupid because they need extremely long time for education. Really, educational time and level of intelligence are two different parameters of the mind and if one of them is better then this can be at the expense of the other. For example, a human being needs about a year to learn to walk, which is much worse in comparison to most animals. Some of the greatest scientists had bad results in school, which can be interpreted as a fact that they advanced slower than the ordinary people.

Therefore, the educational time is important and it has to be limited in order to make our definition useful. This will be done by changing the life length from infinite to finite. We will assume that the length of the life is 100 games. Each game has maximum 1000 steps, which means that the life length is not bigger than 10,000 steps. Now the success of the life will not be the limit of the Success function but the value of this function for the first 100 games.

After this we can look for program which makes a good success in an arbitrary world, but this is not a good idea because the arbitrary world is too unpredictable. Human beings use the assumption that the world is simple and that is why they cope very well in a more simple environment and they are totally confused if the environment is too complicated. Therefore, we have to restrict the complexity of the world and give bigger importance to the more simple worlds. For this restriction we will use Kolmogorov Complexity [8]. The parameter which restricts the complexity of the world will be the level of intelligence of AI.

## Kolmogorov Complexity

First we need a definition of program which calculates the functions World and View. For this we will use the same definition of TM as for the program which was our AI. There will be some small differences: The alphabet of the Turing Machine of the world (TM_W) will be $\Sigma \cup \Omega \cup \{\lambda\}$ (the only service symbol will be $\lambda$). Also, TM_W will input the row $d_1, d_2, d_3, ...$ and output the row $v_1, v_2, v_3, ...$ . At the beginning TM_W will start with tape on which $d_1$ is at the head position and the rest is $\lambda$. At the end of the first big step TM_W will output $v_1$ and input $d_2$. F will be set of 5-tuples which is a subset to $P \times \Delta \times P \times \Delta \times$ {Left, Right}. This means that F is not a function but a relation (because it will represent multy-valued function). We will assume that $\forall s \ \forall \delta \ \exists$ 5-tuple$\in$F whose first two elements are s and $\delta$ (this makes the reasons for hanging with one less). The 5-tuples in F whose third element is $p_0$ will be called output 5-tuples. The fourth element of output 5-tuples has to be letter from $\Sigma$ (this is not necessary but sufficient condition in order TM_W to output only letters from $\Sigma$). We will allow nondeterministic behavior only for output 5-tuples. This means that if two different 5-tuples have the same first and second elements then they both have to be output 5-tuples. There will be no two 5-tuples which differ only at the fifth element (we cannot have a choice between two nondeterministic 5-tuples which output the same letter - look again at remark 3). It will be more interesting if we assume that nondeterministic 5-tuples have additional parameter which shows the possibility for each of them to be chosen. Nevertheless, we will assume that this possibility is distributed equally and that we do not have such additional parameter.

According to this definition, the internal states of the world will be the states of the tape of the TM_W. If we want to have worlds without fatal errors we have to clean the tape of TM_W after each game (after printing any final letter). Nevertheless, we will not do this because the absence of fatal errors was important when we had infinite life and when we counted on that sooner or later all errors will be compensated. For real AI is better to assume some connection between games. Otherwise AI will not remember what was done in the last game or it will remember it but it will not know whether this was in the last game or in some other game from the previous ones.

Another question is what we will do with TM_W which hangs. We do not want to exclude these programs from our definition (at least because we cannot check this characteristic). That is why we will assume that if one TM_W makes more than 800 small steps without making a big step then we will interrupt it with output "draw". This means that it will do the next small step in the same way as if the 5-tuple executed at this moment had third element $p_0$ and fourth element "draw". Also, if one TM_W makes 1000 big steps without outputting any final symbol then the output of the next big step will be "draw". We need this in order to keep the games finite, which is important in order to keep the life finite (the life is 100 games).

We will define the size of TM_W as the number of internal states plus the level of indefiniteness (this is the minimal number of nondeterministic 5-tuples, which have to be deleted from F in order to make it deterministic or this is the number of all nondeterministic 5-tuples minus the number of different groups of nondeterministic 5-tuples).

So, we will restrict the set of possible worlds to those generated by Turing Machine whose size is not bigger than 20. The maximum size of the TM_W will be the level of intelligence of AI. The simpler worlds will be more important because they are generated from more than one TM_W and that is why we will count their result more than once.

**Remark 4**: It looks like that two Turing machines (the world and AI) play against each other. Anyway, this is wrong because the world does not care about AI and it dose not play against AI.

## Final definition of AI

Now everything is fixed. We have finite lives, which are exactly 100 games. We had selected the success function that will evaluate these lives. Also, we made a finite set of worlds which consist of the worlds generated from the TM_W with size not bigger than 20. Now we can define AI as this program which will make the best average success in the selected worlds. Such program exists and it will be the next trivial decision (TD5).

The number of all strategies for playing 100 consecutive games is finite. The number of selected worlds is also finite. We can calculate the expected success of any strategy in any world. The average success of a strategy will be the arithmetical mean from its expected success in any world. (The calculation of the expected success of a strategy in a world is easy if the world is deterministic. In this case, we will have simply to play 100 games with this strategy in this world. In the opposite case, if the world is nondeterministic then we have to use Max-Sum algorithm, which is theoretically possible, but in practice it is a combinatory explosion. Nevertheless, even if the worlds were only deterministic, we would have combinatory explosion again from the number of worlds and from the number of strategies.)

Hence, TD5 will be this program which calculates and plays the best strategy (this which average success is biggest). Such program is easy to be written but it is very difficult to wait until it makes its first step. The situation with the perfect chess playing program is analogical. (It plays chess by calculating all possible moves.) This program can be written very easy but the time until the end of the universe will be not enough for it to make its first move.

It will be too restrictive if we define AI as the best program (such as TD5 or such as any other program equivalent to TD5). It will be better if we say that AI is a program whose average success is not more than 10% from the best (from TD5) (when we say 1% we mean the number 0.01) (the distance to the best is something completely different from the parameter "courage" in TD2). Such definition is theoretically possible, but practically inconvenient. The reason for this is the fact that the value of the average success of TD5 can be theoretically calculated, but in practice this is absolutely impossible. So, if we select such definition we will not be able to check it for a concrete program.

Final definition: **AI will be a program which makes more than 70% average success in the selected set of worlds.**

**Assumption 1**: Here we assume that the average success of TD5 is about 80%. If this conjecture is true then there exists a program which satisfies the definition (at least TD5 do). If the average success of TD5 is smaller than 70% then there is no such a program (of course, in such case we can change this parameter and make it smaller than 70%).

The advantage of this definition of AI is that we can check it for a concrete program. Of course, we cannot calculate the average success for any program due to the combinatory explosion, but we can calculate it approximately by the methods of the statistics. How we will do this. We will select at random N worlds (world is TM_W with size not bigger than 20) and we will play 100 consecutive games in every world. If the world is deterministic then this will give us the expected success of our program in this world. If it is not deterministic then we will play at random in this world. This will give us statistically good evaluation of the expected success because the possibility to be extremely lucky in 100 games is very small (so is the possibility to be extremely unlucky). If N (the number of the tested worlds) is big then the statistical result will be close to the average success of our program.

If $|\Sigma \cup \Omega \cup \{\lambda\}| = 5$ (which is the minimum - remark 2) then the number of deterministic TM_W with 20 states is 200 on power of 100. If we take the number of nondeterministic TM_W with 19 states and level of indefiniteness one (which means with two nondeterministic 5-tuples) then this number is many times smaller than 200 on power of 100. In order to use the method of statistics we have to calculate how many times is this number smaller. Otherwise we will use wrong correlation between deterministic and nondeterministic TM_W. Anyway, such wrong correlation will make an unessential change in the definition of AI.

## Conclusion

Now we have definition of AI and at least one program (TD5) which satisfies it (with assumption 1). The first question is: Does this definition satisfy our intuitive idea that AI is a program which is more intelligent than a human being. Yes, but for some values of the parameters educational time and level of intelligence. In this paper the educational time was fixed on 100 games each of them no longer than 1000 steps (educational time is equal to the life length because we learn all our life). The level of intelligence here was fixed on 20. Which means that we assume that we can find a model of the world which is TM_W with size not bigger than 20. We cannot say what is the exact level of intelligence of the human being.

The second question is: Is TD5 which satisfies the definition the program which we are looking for. The answer is definitely no. We are looking for a program which can work in real time. Also, our intuitive idea is that AI should build a model of the world and on the base of it AI should plan its behavior. Look at [6, 7]. Instead of this, TD5 uses brutal force in order to find the best strategy. Even TD5 will not know what to do on the game 101 because its strategy is only for 100 games.

Here we will offer the last trivial decision (TD6), which corresponds better to our intuitive idea for AI.

Let TD6 be the program which tries all deterministic TM_W and accepts as a model of the world the first one (the shortest one) which generates the life until the present moment. After selecting a model (which will be a big problem for more complicated worlds) TD6 will use this model and the Max-Sum algorithm in order to plan its next move. Here the Max-Sum algorithm has two modifications: First, there is no Sum (or there is only one term in the sum) because the modeling TM_W is deterministic and there is only one possible reaction from the world. Second modification is that Max-Sum will not calculate until the end of the life or even until the end of the game because this will give a combinatory explosion. Instead, it will calculate several steps like chess playing programs do.

You can find a program similar to TD6 in [5]. Really, in [5] we are looking for TM_W, which is a generator of infinite sequence $v_1, v_2, v_3, ...$ instead of looking for transducer from $d_1, d_2, d_3, ...$ to $v_1, v_2, v_3, ...$ . Also, [5] does not make moves (there is no Max-Sum algorithm for calculating the next move). The only thing which [5] does is to predict the next number of the sequence. Anyway, in [5] you can see that searching for a model in the set of all TM_W works only if the model is very simple. If the size of TM_W is bigger than 5, the result is combinatory explosion.

If we modify TD6 in order to accept also the nondeterministic Turing machines as models of the world then we will have too much possible models. In this case we have to consider more than one model and to evaluate each possible model in order to see how reliable it is.

Anyway, TD6 and its modifications are not the program which we are looking for, although TD6 can be done to satisfy the definition (because the definition does not say anything about the efficiency of AI). The program which we are looking for is much closer to that one which is described in [6, 7]. The problem in TD6 is that it looks for a model of the world which consists from only one item. It is better if the model is a set of many items (the items can be Turing machines, final automata or logical formulas). When we make a theory in logic then it consist from a set of axioms and we can change smoothly the theory by modifying, adding or deleting one axiom. Any theory in logic is a model of some world. AI has to use similar models which can be modified smoothly.